\documentclass[runningheads]{llncs}
\usepackage[T1]{fontenc}
\usepackage{graphicx}
\usepackage{makeidx}  
\usepackage[colorlinks,linkcolor=blue]{hyperref}
\usepackage{url}
\usepackage{longtable}
\usepackage{multirow}
\usepackage{amssymb, amsmath, bm}
\usepackage{amsfonts, amssymb}
\usepackage{pifont}
\usepackage{mathtools}
\usepackage{mathrsfs}
\usepackage{booktabs}
\usepackage{cite}
\usepackage{dsfont}
\usepackage{makecell}
\usepackage[noend]{algpseudocode}
\usepackage{algorithmicx,algorithm}
\usepackage[misc]{ifsym} 
\usepackage{bbding}
\usepackage{ulem}
\usepackage[utf8]{inputenc}
\newcommand{\M}{HeartBeat}

\begin{document}

\title{HeartBeat: Towards Controllable Echocardiography Video Synthesis with Multimodal Conditions-Guided Diffusion Models}

\titlerunning{\M}

\author{Xinrui Zhou\inst{1,2,3}\thanks{Xinrui Zhou and Yuhao Huang contribute equally to this work.}, Yuhao Huang\inst{1,2,3\star} \and Wufeng Xue\inst{1,2,3} \and Haoran Dou\inst{4} \and \\Jun Cheng\inst{1,2,3} \and Han Zhou\inst{1,2,3} \and Dong Ni\inst{1,2,3}\textsuperscript{(\Letter)}} 


\institute{
\textsuperscript{$1$}National-Regional Key Technology Engineering Laboratory for Medical Ultrasound, School of Biomedical Engineering, Medical School, Shenzhen University, China\\
\email{nidong@szu.edu.cn} \\
\textsuperscript{$2$}Medical Ultrasound Image Computing (MUSIC) Lab, Shenzhen University, China\\
\textsuperscript{$3$}Marshall Laboratory of Biomedical Engineering, Shenzhen University, China\\
\textsuperscript{$4$}Centre for Computational Imaging and Simulation Technologies in Biomedicine (CISTIB), University of Leeds, UK\\}

\authorrunning{X. Zhou and Y. Huang et al.}
\maketitle              

\begin{abstract}
Echocardiography (ECHO) video is widely used for cardiac examination. 
In clinical, this procedure heavily relies on operator experience, which needs years of training and maybe the assistance of deep learning-based systems for enhanced accuracy and efficiency.
However, it is challenging since acquiring sufficient customized data (e.g., abnormal cases) for novice training and deep model development is clinically unrealistic.
Hence, controllable ECHO video synthesis is highly desirable. 
In this paper, we propose a novel diffusion-based framework named \M~towards controllable and high-fidelity ECHO video synthesis.
Our highlight is three-fold. 
First, \M~serves as a unified framework that enables perceiving multimodal conditions simultaneously to guide controllable generation.
Second, we factorize the multimodal conditions into local and global ones, with two insertion strategies separately provided fine- and coarse-grained controls in a composable and flexible manner. 
In this way, users can synthesize ECHO videos that conform to their mental imagery by combining multimodal control signals.
Third, we propose to decouple the visual concepts and temporal dynamics learning using a two-stage training scheme for simplifying the model training.
One more interesting thing is that \M~can easily generalize to mask-guided cardiac MRI synthesis in a few shots, showcasing its scalability to broader applications.
Extensive experiments on two public datasets show the efficacy of the proposed \M.
\end{abstract}

\section{Introduction}
Cardiac ultrasound (US), i.e., echocardiography (ECHO), is widely utilized to evaluate cardiac function and diagnose cardiovascular diseases. 
Compared to other modalities, US has the advantages of real-time imaging, radiation-free, and affordability in clinical practice~\cite{wei2020temporal}. 
ECHO is a dynamic examination that heavily relies on operator experience. It usually takes more than 10 years to train qualified sonographers. 
Recent deep learning-based systems have shown efficacy in automatic cardiac diagnosis, which can assist sonographers in clinical analysis~\cite{ouyang2020video}.
Collecting abundant ECHO videos tailored to specific anatomy composition (e.g., plane types, specific mitral valve (MV) motion directions, etc.) or abnormal cases has the potential to accelerate novice training and intelligent system development.
However, collecting adequate customized ECHO sequences is impractical in clinical scenarios~\cite{liang2022weakly}.
Hence, building a framework for controllable ECHO video synthesis is greatly desired to solve the data scarcity issue.

ECHO video synthesis is challenging due to the US speckle noise, complex motion trajectories (e.g., mitral valve motion), and varying sizes of anatomical structures.
Recently, driven by the significant progress of the Denoising Diffusion Probabilistic Models (DDPMs)~\cite{ho2020denoising}, several DDPM-based approaches have been proposed to synthesize ECHO videos. 
Stojanovski et al.~\cite{stojanovski2023echo} synthesized 2D ECHO videos with the guidance of semantic masks. 
Though generating realistic images, it focused on static image synthesis instead of dynamic sequences, limiting its clinical practice. 
Reynaud et al.~\cite{reynaud2023feature} proposed a cascaded video diffusion model to generate ECHO videos conditioned on the end-diastolic frame and left ventricle ejection fractions. 
Van et al.~\cite{van2024echocardiography} introduced a single semantic mask condition to compose ECHO videos. 
However, most of these methods show poor controllability and flexibility due to restricted control signals. 
Moreover, they directly utilized 3D diffusion architectures, resulting in substantial computational cost and training difficulty. 
Thus, they may not be suitable for clinical practice.

Most recently, studies have been proposed to achieve controllable video synthesis. 
They can be coarsely classified into two types. 
\textbf{(1) Text-to-Video (T2V) Synthesis.} 
Most approaches~\cite{blattmann2023align, ho2022imagen, singer2022make, zhou2022magicvideo} took the random Gaussian noise and text prompt as inputs and learned both visual concepts and temporal dynamics. 
However, these methods typically fell short in controlling the visual appearance and geometric structure of the generated videos due to the lack of fine-grained conditional controls. 
\textbf{(2) Multiple Conditions-guided T2V Synthesis.} 
Compared to pure T2V synthesis, these methods enabled controllable and precise synthesis driven by multiple conditions. 
For facilitating the global controls regarding visual appearance, AnimateDiff~\cite{guo2023animatediff} and MoonShot~\cite{zhang2024moonshot} conditioned the synthesis on both image and text inputs simultaneously. 
However, these methods overlooked fine-grained controls, limiting their applicability in the medical field. 
For local controls related to geometric structure, ControlVideo~\cite{zhang2023controlvideo} reused the pretrained ControlNet~\cite{zhang2023adding} that was designed for conditional image generation to achieve controllable video synthesis. 
Wang et al.~\cite{wang2024videocomposer} proposed VideoComposer to achieve multiple conditions-guided T2V synthesis in a composable fashion. 
However, it required high training costs by introducing numerous temporal layers to model the time-series knowledge of conditions.

In this study, we propose a novel framework called \M~to achieve controllable and high-fidelity ECHO video synthesis. 
We believe that this is the first exploration of highly customized US video synthesis based on the guidance of multimodal conditions.
Our contributions are three-fold. 
First, we introduce \M, a uniform framework that enables perceiving versatile conditions simultaneously to guide controllable video generation.
Second, we factorize the multimodal conditions into local (i.e., structural cues like sketches, hand-crafted motion direction of MV, etc.) and global (i.e., plane types and image priors) parts.
For instance, sketches allow for flexible edits in anatomical structures and hand-crafted strokes of MV make its motion controllable. 
Moreover, generating abnormal cases becomes feasible due to the transfer of visual patterns in image priors.
Two corresponding insertion methods are further proposed to provide fine- and coarse-grained controls, respectively. 
Note that all conditions can be manipulated in a composable and flexible manner. 
In this case, users can seamlessly combine multimodal conditions to produce ECHO videos. 
Third, we propose a two-stage training scheme that decouples the visual concepts and temporal dynamics learning to ease the model training. 
Furthermore, we validate the generalization ability of \M~by transferring ECHO video synthesis into 3D cardiac MRI (CMR) generation using few-shot learning. 
Extensive experimental results prove the effectiveness of \M.

\section{Methodology}
Fig.~\ref{fig:framework} shows the pipeline of our proposed~\M.
It supports composable and customized ECHO video synthesis via multimodal conditions.
\M~involves a two-stage training scheme design. 
In the pretraining stage, \M~focuses on high-quality \textit{visual concepts learning} for controllable text-to-image (T2I) generation. 
While in the finetuning stage, the domain-specific knowledge gained from the pretext task is reused to facilitate \textit{temporal dynamics modeling} for customized text-to-video (T2V) generation. 
During inference, given a group of multimodal prompts as conditional inputs, \M~enables customizable ECHO video synthesis from Gaussian noise.

\subsection{Preliminaries of T2V DDPMs.}
T2V DDPMs are generative models developed upon traditional DDPMs~\cite{sohl2015deep, ho2020denoising}. 
They are trained to learn video distribution by iteratively recovering noisy inputs with the guidance of text prompts. 
To reduce the computational burden caused by pixel-space training, latent diffusion models (LDMs) that operate in the latent space are introduced in the video domain for perceptual video compression. In this way, the optimized objective is adapted to 
$\min_\theta \mathbb{E}_{\mathit{z}_0,\epsilon, \mathbf{c}, \mathit{t}}[||\epsilon - \epsilon_\theta(\mathit{z}_t, \mathbf{c}, \mathit{t})||_2]$, where $\mathit{z}_0$ is the latent representation of the training video from a pretrained variational autoencoder (VAE)~\cite{esser2021taming}.
\textit{t} stands for the time step. 
$\epsilon_\theta$ and $\epsilon$ represent the predicted and target noise, respectively.
For our model, $\mathbf{c}$ refers to multimodal conditions including local and global conditions.

\subsection{\M}
\subsubsection{Decoupled learning of visual concepts and temporal dynamics.} 
Directly employing video LDMs (VLDMs) is an intuitive way to achieve ECHO video synthesis. 
However, it is challenging to simultaneously model visual appearance and temporal variation driven by scarce medical data, while producing high training costs. 
Thus, in \M, we first pretrain an LDM and extend it to a VLDM for controllable video generation. 
Note that the VLDM is an inflation of the 2D counterpart over the spatial-temporal dimension. 
Leveraging the decoupling training scheme, \M~can achieve high-fidelity and temporal-coherent ECHO video synthesis, while reducing training difficulty and computational burden.

\begin{figure*}[!t]
    \centering
    \includegraphics[width=1.0\linewidth]{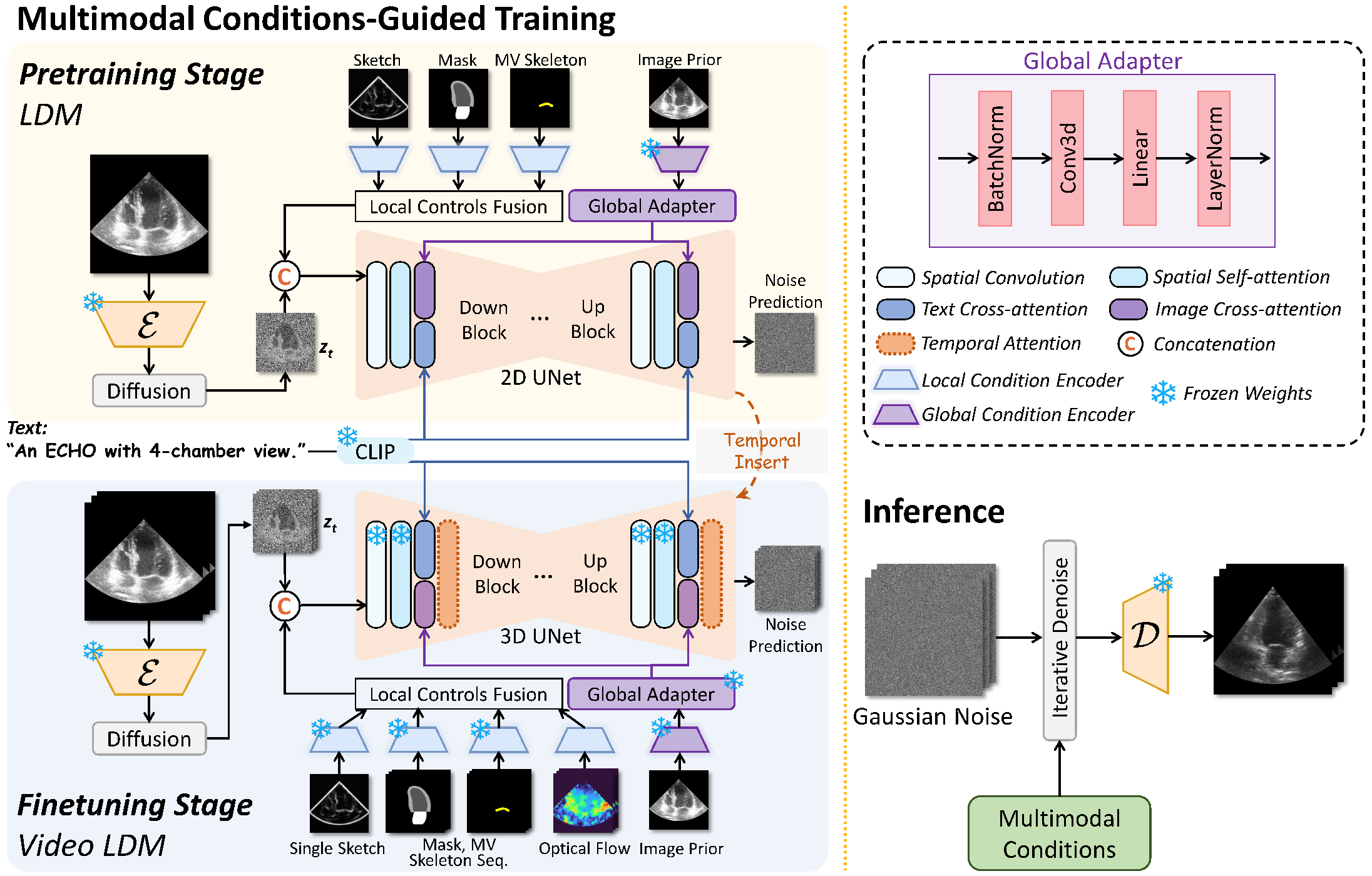}
    \caption{Pipeline of \M. $\mathcal{E}, \mathcal{D}$ denote pretrained encoder and decoder in VAE, respectively. $z_t$ refers to latent features. LDM, latent diffusion model. MV, mitral valve.} 
    \label{fig:framework}
\end{figure*}

\subsubsection{2D-to-3D Model Inflation for Video Synthesis.} 
In the pretraining stage, following LDM~\cite{rombach2022high}, we employ a 2D UNet~\cite{ronneberger2015u} as the denoising network for controllable image generation. 
Typical UNet in LDMs is composed of stacked blocks, where each contains a spatial convolution layer, a spatial self-attention layer, and a cross-attention layer that controls the synthesis by texts. 
Motivated by~\cite{ho2022video, wu2023tune}, we extend the 2D UNet to the 3D counterpart using a simple inflation strategy. 
Specifically, we inflate all the spatial convolution layers at the temporal dimension with \textit{t=1}. 
To model the temporal dynamics, we insert a temporal self-attention layer following the cross-attention layer in each block. 
This design ensures the feature distribution of the spatial layers will not be altered significantly~\cite{zhang2024moonshot}.
Thus, \M~can reuse the rich visual concepts regarding ECHO video patterns preserved in LDM and focus on temporal features integration.

\subsubsection{Controllable Video Synthesis with Multimodal Conditions.} 
It is well acknowledged that pure T2I/T2V models suffer from limited control of spatial composition~\cite{zhang2023adding}. 
Generating customized ECHO videos of normal and abnormal cases that precisely match their mental imagery is crucial for junior sonographers to improve their reading and diagnostic skills. 
To achieve this, our proposed \M~introduces multimodal conditions to guide customizable ECHO video synthesis in a flexible and composable manner.

As shown in Fig.~\ref{fig:framework}, the overall generation process is comprehensively controlled by employing multimodal control signals. 
In this case, \M~allows users to flexibly select/edit any available single condition or their combination to guide high-fidelity, temporal-coherent, and customized ECHO video synthesis. 
In our study, we involve six multimodal conditions and factorize them into four local conditions and two global conditions for fine-grained and coarse-grained controls, respectively.
To fully perceive multimodal conditions at local and global dimensions, we separately design condition injection methods for two conditions. 
Meanwhile, \M~assures the composability of different conditions by adopting such injection strategies. 
Note that LDM and VLDM in the proposed \M~share identical conditions injection methods.

\textbf{1) Local Conditions Guidance for Fine-grained Control.} 
To achieve fine-grained geometric structure guidance, we apply \textit{sketch}, \textit{semantic mask}, and \textit{skeleton of mitral valve} as three local spatial conditions to perform controllable image generation. 
To improve the temporal consistency of generated videos, in VLDM, we extend the above local spatial conditions to time-series ones except for \textit{sketch}. 
Optical flow is also employed as an additional local condition to explicitly indicate the pixel-wise movements between adjacent frames (see Fig.~\ref{fig:framework}).
We propose a \textbf{local condition injection method}, which consists of a lightweight encoder~\cite{esser2021taming} for embedding each local condition and a fusion operation between conditions and noisy latents $\mathit{z_t}$. 
Specifically, each condition is first encoded in parallel, allowing \M~to capture local spatial knowledge. 
The obtained local conditional features are then fused by element-wise addition. 
Last, such features are concatenated with $\mathit{z_t}$ along channel dimension to form local control signals. 
Note that each encoder shares the same architecture. 

\textbf{2) Global Conditions Perception for Coarse-grained Control.} 
Towards highly-customized video synthesis solely guided by local conditions has a limitation, that is, the synthetic results lack diversity due to the various fine-grained conditional inputs. 
However, acquiring sufficient diverse ECHO videos is desirable for novices undergoing clinical training or for deep learning systems development.
Thus, we also consider two global conditions, i.e., \textit{text} and \textit{image prior} to balance the trade-off between highly-customized and diverse generations (see Fig.~\ref{fig:framework}). 
The text offers an intuitive indication in terms of coarse-grained visual content, while the latter provides richer information beyond text (e.g., visual patterns regarding abnormal disease).
We develop a \textbf{global condition injection method} to align the global image prior embeddings $\mathit{f_i}$ extracted from the pretrained MedSAM image encoder~\cite{huang2024segment} with the text embeddings $\mathit{f_t}$ derived from CLIP text encoder~\cite{radford2021learning}. 
For the global conditions integration, an intuitive way is to directly input concatenated features of both into the frozen cross-attention layers. 
However, it will hinder the model from capturing visual patterns from the image prior. 
To this end, we replace the original text-guided cross-attention layer in each UNet block with a factorized cross-attention layer for handling texts and image priors. 
In this layer, two separate Query-Key-Value (QKV) projections are added and optimized for both conditions, respectively. 
Take the proposed VLDM as an example, this attention process can be defined as: $\mathit{CrossAttention}(\mathbf{Q}^T,\mathbf{K}^T, \mathbf{V}^T) + \mathit{CrossAttention}(\mathbf{Q}^I,\mathbf{K}^I, \mathbf{V}^I)$, where $\mathit{Q}^T, \mathit{Q}^I \in \mathbb{R}^{BF \times H \times W \times C}$, $\mathit{K}^T, \mathit{K}^I, \mathit{V}^T, \mathit{V}^I \in \mathbb{R}^{BF \times N \times C}$, with $\mathit{B}$ the batchsize, $\mathit{F}$ the length of frames, $\mathit{H}$ the height, $\mathit{W}$ the width, $\mathit{N}$ the token numbers of text ($\mathit{T}$) and image prior condition ($\mathit{I}$), and $\mathit{C}$ the hidden dimension. 
Besides, we propose a learnable global adapter to align $\mathit{f_i}$ and $\mathit{f_t}$. 

\section{Experimental Results}
\begin{table}[!t]
    \centering
    \scriptsize
    \caption{Quantitative results of methods. Conditional controls involve text (T), sketch (S), image prior (I), mask (M), the skeleton of the mitral valve (MV), and optical flow (O). Note that "\M" and "VideoComposer" use the joint training strategy, while other models merely use selected conditions for training. }
    \resizebox{\textwidth}{!}{
    \begin{tabular}{cccccccccccccc}
    \toprule
    \multirow{2}[2]{*}{Methods} & \multicolumn{6}{c}{Controls} & \multicolumn{3}{c}{A2C} & \multicolumn{3}{c}{A4C} \\
    \cmidrule(lr){2-7} \cmidrule(lr){8-10} \cmidrule(lr){11-13}
    & T & S & I & M & MV & O & FID$\downarrow$ & FVD$\downarrow$ & SSIM$\uparrow$ & FID$\downarrow$ & FVD$\downarrow$ & SSIM$\uparrow$ \\
    \midrule
    MoonShot~\cite{zhang2024moonshot} & \checkmark &  & \checkmark &  &  &  & 48.44 &                                   12.65 & 0.63 & 61.57 & 18.13 & 0.62 \\ 
    VideoComposer~\cite{wang2024videocomposer} & \checkmark & \checkmark &      
                                               \checkmark & \checkmark & \checkmark & \checkmark & 37.68 & 
                                               10.31 & 0.60 & 35.04 & 11.27 & 0.61 \\
    \midrule
    \M-3D & \checkmark &  &  & \checkmark &  &  & 54.63 & 13.64 & 0.60 & 41.07 & 14.01 & 0.60 \\ 
    \M-Base & \checkmark &  &  & \checkmark &  &  & 23.57 & 6.97 & 0.65 & 31.08 & 10.80 & 0.64 \\
    \midrule
    \multirow{6}{*}{\M} & \checkmark &  &  &  &  &  & 107.66 & 19.07 & 0.53 &                                   76.46 & 23.83 & 0.53 & \\
                        & \checkmark &  & \checkmark &  &  &  & 36.34 & 9.60 & 0.61 & 36.00 & 12.58 & 0.61 & \\
                        & \checkmark & \checkmark & \checkmark &  &  &  & 27.38 & 7.04 & 0.66 & 35.42 & 13.48 & 0.64 &\\
                        & \checkmark & \checkmark & \checkmark & \checkmark &  &  & 26.30 & 6.74 & 0.66 & 33.77 & 12.59 & 0.64 &\\
                        & \checkmark & \checkmark & \checkmark & \checkmark & \checkmark &  & 25.98 & 6.94 & 0.66 & 33.98 & 12.00 & 0.64 &\\
                        & \checkmark & \checkmark & \checkmark & \checkmark & \checkmark & \checkmark & \textcolor{blue}{25.23} & \textcolor{blue}{6.08} & \textcolor{blue}{0.66} & \textcolor{blue}{31.99} & \textcolor{blue}{9.96} & \textcolor{blue}{0.65} &\\
    \bottomrule
    \end{tabular}}
\label{tab:results}
\end{table}

\subsubsection{Dataset and Implementations.} 
We validated \M~on the public CAMUS dataset~\cite{leclerc2019deep}. 
It consists of 900 ECHO videos collected from 450 patients. 
Videos labeled with poor quality in~\cite{leclerc2019deep} were excluded for reliability. 
Finally, 884 videos with 431 apical two-chamber (A2C) and 453 apical four-chamber (A4C) views were included. 
The dataset was split randomly into 793 and 91 videos with 16 frames for training and testing at the patient level. 
We set text prompts (i.e., "An ECHO with 2/4-chamber view.") for all videos according to the actual view. 
Sketches were extracted by PiDiNet~\cite{su2021pixel}. 
For few-shot generalization validation, we employed 50 CMR volumes from M$\&$Ms Challenge~\cite{campello2021multi} as the training set. 

\begin{figure*}[!t]
    \centering
    \includegraphics[width=1.0\linewidth]{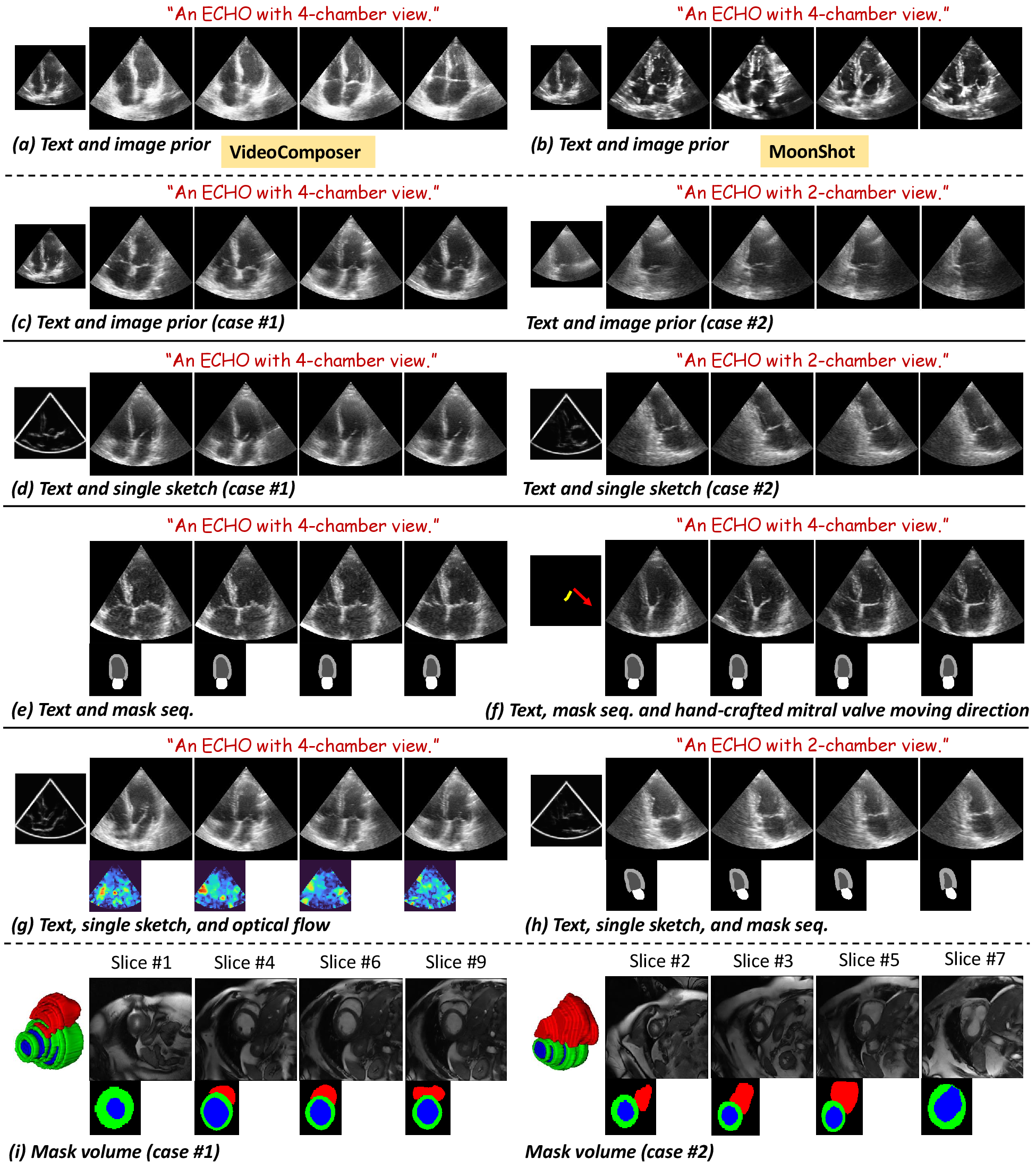}
    \caption{Typical results of \M~and two most related approaches (first row).} 
    \label{fig:visualization}
\end{figure*}

In this study, we implemented \M~in \textit{Pytorch}, using four NVIDIA A6000 GPUs. 
All frames/slices were resized to 256$\times$256. 
During pretraining, our LDM was developed upon Stable Diffusion~\cite{rombach2022high} and initialized using the public pretrained weights\footnote{\url{https://huggingface.co/CompVis/stable-diffusion-v1-4}}. 
During finetuning, we froze the spatial weights except for the newly added optical flow encoder and kept the temporal layers trainable. 
We updated the query projection in cross-attention layers to refine the text- and image prior-video alignment. 
We set the batch size as 64 and 16 for the first and second stage training. 
During the whole training, we adopted the all conditions joint training strategy~\cite{huang2023composer}.
In this way, \M~was not required to be finetuned for each unique combination of multimodal conditions every time and enables flexibly dropping several conditions during inference.
The CMR synthesizer was merely conditioned on mask volumes, and initialized with US-trained \M.
The learning rate was initialized to 1e-4 after 500 steps of warm-up strategy and decayed by a cosine annealing scheduler. 
All models were trained using the Adam optimizer for 200 epochs. We chose models of the last epoch to work with \M.

\subsubsection{Quantitative and Qualitative Analysis.}
We validated the generative performance of \M~on achieving various tasks in a controllable manner. 
Conditions can be tailored to meet the specific needs of users by leveraging the composable control capabilities of \M. 
In this study, three metrics were used to evaluate the performance, including 1) Fr\'{e}chet Inception Distance (FID)~\cite{heusel2017gans} for image quality evaluation in feature level, 2) Fr\'{e}chet Video Distance (FVD)~\cite{unterthiner2018towards} for visual quality and temporal consistency assessment in video level, and 3) Structure Similarity Index (SSIM) score~\cite{wang2004image} to assess the controllability~\cite{zhao2024uni}. 

Fig.~\ref{fig:visualization} shows the qualitative results: 
\textbf{1) Image prior-controlled ECHO video synthesis.} 
Fig.~\ref{fig:visualization}(c) shows synthesizing realistic videos driven by global knowledge (i.e., coarse-grained visual patterns) of image priors. 
Given priors with normal (\textit{left}) and left ventricular hypertrophy (\textit{right}), \M~generates videos with the same disease status accordingly, allowing the normal/abnormal transfer abilities of the model.
\textbf{2) Sketch-controlled ECHO video synthesis.} 
\M~is able to animate static sketch for generating realistic videos (Fig.~\ref{fig:visualization}(d)). 
\textbf{3) Mitral valve motion-controlled ECHO video synthesis.} 
Observation of MV motion is clinically important. 
With the simple MV skeleton, masks, and hand-crafted strokes indicating the motion direction, i.e., red arrow in Fig.~\ref{fig:visualization}(f), \M~enables precise motion control compared to the same case with only mask condition (Fig.~\ref{fig:visualization}(e)). 
Such features endow users with ease of use, flexibility, and high controllability when using \M, showcasing the potential for clinical applications. 
\textbf{4) Various conditions-controlled ECHO video synthesis.} 
Fig.~\ref{fig:visualization}(g-h) demonstrates the superior controllability and generative quality of \M.  
\textbf{5) Generalize to 3D CMR synthesis.} 
Thanks to the flexibility of control signals manipulation, \M~enables controllable CMR synthesis with few-shot tuning solely guided by the 3D mask (Fig.~\ref{fig:visualization}(i)). 
It shows that the generated anatomy areas are highly consistent with the masks. 
Besides, \M~produces sharper frames with high fidelity and coherence than others (Fig.~\ref{fig:visualization}(a-c)). 
The quantitative results in Table~\ref{tab:results} are in line with the qualitative ones. 
Notably, the "\M-Base" model with a two-stage training scheme performs better than that with pure 3D one-stage training using  VLDM (i.e., \M-3D), showcasing the efficacy of our training scheme. 
Ablation studies (last 6 rows) show that adding more conditions enhances generative performance. 

\section{Conclusion}
In this study, we propose a novel diffusion-based \M~framework for controllable and flexible ECHO video synthesis. 
Specifically, \M~is driven by multimodal control signals in a composable fashion, including local and global conditions to separately provide fine- and coarse-grained guidance. 
Extensive experiments on two datasets validate the powerful controllability and generality of \M, showing its clinical practicality. 
In the future, we will extend \M~to more challenging datasets for further generality validation.



\bibliographystyle{splncs04}
\bibliography{Paper-1453}

\end{document}